%% file: root.tex
\documentclass[letterpaper, 10 pt, conference]{ieeeconf}  

\IEEEoverridecommandlockouts                              

\usepackage{amsmath}
\usepackage{graphicx}
\usepackage[hyphens]{url}
\usepackage{multirow}
\usepackage{subcaption}
\usepackage{cite}
\usepackage{tabularx}
\usepackage{booktabs}
\usepackage{float}
\usepackage{placeins}

\title{\LARGE \bf
Robust Plant Disease Diagnosis with Few Target-Domain Samples
}

\author{Takafumi Nogami$^{1}$
, Satoshi Kagiwada$^{2}$ 
and Hitoshi Iyatomi$^{1}$
\thanks{$^{1}$Takafumi Nogami and Hitoshi Iyatomi are with the Department of Applied Informatics, Graduate School of Science and Engineering,
        Hosei University, Tokyo, Japan. Emails: takafumi.nogami.7n@stu.hosei.ac.jp, iyatomi@hosei.ac.jp}
\thanks{$^{2}$Satoshi Kagiwada is with the Department of Clinical Plant Science, Faculty of Bioscience and Applied Chemistry,
        Hosei University, Tokyo, Japan. Email: kagiwada@hosei.ac.jp}
}

\begin{document}

\maketitle
\thispagestyle{empty}
\pagestyle{empty}

\begin{abstract}
Various deep learning-based systems have been proposed for accurate and convenient plant disease diagnosis, achieving impressive performance.
However, recent studies show that these systems often fail to maintain diagnostic accuracy on images captured under different conditions from the training environment---an essential criterion for model robustness.
Many deep learning methods have shown high accuracy in plant disease diagnosis. However, they often struggle to generalize to images taken in conditions that differ from the training setting. This drop in performance stems from the subtle variability of disease symptoms and domain gaps---differences in image context and environment.
The root cause is the limited diversity of training data relative to task complexity, making even advanced models vulnerable in unseen domains.
To tackle this challenge, we propose a simple yet highly adaptable learning framework called Target-Aware Metric Learning with Prioritized Sampling (TMPS), grounded in metric learning.
TMPS operates under the assumption of access to a limited number of labeled samples from the target (deployment) domain and leverages these samples effectively to improve diagnostic robustness.
We assess TMPS on a large-scale automated plant disease diagnostic task using a dataset comprising 223,073 leaf images sourced from 23 agricultural fields, spanning 21 diseases and healthy instances across three crop species.
By incorporating just 10 target domain samples per disease into training, TMPS surpasses models trained using the same combined source and target samples, and those fine-tuned with these target samples after pre-training on source data. It achieves average macro F1 score improvements of 7.3 and 3.6 points, respectively, and a remarkable 18.7 and 17.1 point improvement over the baseline and conventional metric learning.

\end{abstract}

\section{INTRODUCTION}

Damage caused by plant diseases to crops has become a serious problem~\cite{sastry2014management}. However, experts typically perform diagnoses through visual judgment, which has raised concerns about its availability and cost. 
In recent years, many plant disease diagnosis systems---especially those based on convolutional neural networks (CNNs)---have been proposed to reduce diagnosis costs due to their many advantages, such as easy learning and very high discriminative power, which have been reported~\cite{Toda2019, Saleem20201, Fuentes2021, Mohanty2016, guerrero2023monitoring}. However, a critical problem has been pointed out: the diagnostic accuracy is significantly degraded for data from a shooting environment different from the training dataset~\cite{Mohanty2016, Ferentinos2018311, shibuya2022validation}.

In each field where they were taken (i.e., domain), plant images have similarities in terms of variety, background, composition, photographic equipment, and how disease symptoms appear. These commonalities result in substantial differences in image characteristics from one field to another. Shibuya et al. analyzed over 220,000 images of multiple crops captured under real-field conditions. Their study showed that when both training and test images came from the same field, macro F1 scores indicated a discrimination performance of 98.2\% to 99.5\%~\cite{shibuya2022validation}. In contrast, when test images originated from different fields, the performance dropped significantly, with macro F1 scores ranging from 49.6\% to 87.6\%. They reported that even with many high-resolution images, significant differences in image characteristics between fields, known as domain gaps, make it challenging to maintain diagnostic performance in unseen fields. For many plant diseases, notable symptom areas occupy only a tiny proportion of the image and exhibit significant diversity. Therefore, the discriminator adapts to domain-specific features rather than relying on disease symptoms as diagnostic cues, resulting in overfitting and poor diagnostic performance on data from imaging environments different from those used during training. In commonly used data partitioning methods such as hold-out and cross-validation, potentially highly similar data is acquired in the same shooting environment, which increases the similarity between the training and test data. Thus, although the apparent diagnostic performance may be high, the model's actual performance must be tested on entirely unseen data from fields different from the training data~\cite{shibuya2022validation, guth2023lab, Wayama2024}. This highlights significant room for improvement in developing practical diagnostic systems. 

The underlying factor causing these problems due to domain gaps is the lack of diversity in the training data. However, collecting sufficiently diverse training data to cover the data distribution of unseen fields remains challenging.
As a countermeasure, in addition to various data augmentation methods, one common approach is to suppress the influence of background regions, which often reflect field-specific characteristics. For instance, extracting regions of interest (ROI) that include disease-relevant areas such as leaves has shown promising results~\cite{Saikawa20195177}. Nevertheless, recent studies suggest that even within ROI, domain-specific biases can persist, limiting the effectiveness of background suppression alone~\cite{shibuya2022validation}.

Several methods have been proposed to compensate for the lack of diversity in training data by generating new training data using GAN-based generative models, and some success has been achieved~\cite{guerrero2023monitoring,Cap20221258,Kanno2021554,Arsenovic2019}. Due to the limited diversity of the generated images, it is not possible to compensate for enough diversity to cover unseen fields when the domain gaps are significant. Generative methods based on the Latent Diffusion Model~\cite{Rombach202210674} are still in their early stages, with several techniques aiming to resolve domain gaps~\cite{Dunlap2023,hemati2023cross,dunlap2023diversify,zang2023boosting}. Nonetheless, they have yet to be applied to plant disease diagnosis, and future development is desirable.

On the other hand, transfer learning methods called domain adaptation have achieved excellent results in addressing large domain gaps where data from the target domain can be partially observed~\cite{Fuentes2021,Ganin2017189,Tzeng20172962,Saito20183723,Wu2023,Yan20232555}. Domain adaptation applies knowledge obtained from a domain with sufficient data (i.e., source domain) to a related domain (i.e., target domain) to improve the performance of the discriminator in the target domain. When labeling in the target domain is challenging, unsupervised domain adaptation (UDA), which uses large amounts of labeled data from the source domain and unlabeled data from the target domain, is used. Although UDA has been applied to plant disease diagnosis with some success~\cite{Fuentes2021, Wu2023, Yan20232555}, it still faces notable challenges. In particular, its effectiveness tends to decline when there is a significant distribution shift between the source and target domains. Moreover, the adaptation process typically requires a large volume of unlabeled data from the target domain, which may not always be readily available.
If a large amount of labeled data in the target domain were accessible, one would expect to create a highly accurate model. However, obtaining such data is expensive, which is not a desirable scenario. A small amount of labeled data in the target domain will likely be available when implementing a diagnostic model. Hence, setting up a small amount of labeled data to be observable is a realistic next-best option.

Therefore, we carefully consider feasibility as a realistic means of constructing a high-performance diagnostic system. This study finds measures to relax the problem's constraints to allow a minimal number of labeled images from the target field and use the information to its fullest extent. To maximize little target domain information under these conditions, we propose Target-Aware Metric Learning with Prioritized Sampling (TMPS), a learning method that adapts the model to the target domain, adopting the idea of metric learning. 
TMPS is a straightforward and versatile learning strategy that offers a promising solution for tasks with large domain gaps in general. This approach can be particularly effective for tasks where the cost of obtaining labeled data is extremely high, such as in medical data, or where within-class diversity is vast, as in plant disease diagnosis, making simple domain adaptation challenging. 
In this report, we evaluate the effectiveness of TMPS in the automatic diagnosis of plant diseases, one of the applications for which a solution is particularly needed. We conducted experiments using 223,073 leaf surface images of plants taken in a real field consisting of 3 crops, 21 diseases, and health.

\section{Related Work}

\subsection{Conventional Data Augmentation}
Data augmentation is a technique used to artificially increase the diversity of limited training data by applying random transformations such as image rotation, brightness adjustment, and noise injection. A wide variety of augmentation methods exist, many of which are cost-effective. This approach is widely adopted as one of the most common and effective strategies to improve generalization in machine learning. This is particularly important in the agricultural field, where collecting large-scale datasets of disease images can be challenging~\cite{ArunPandian2019199}.

\subsection{Region of interest (ROI)}
Fujita et al.\cite{Fujita201849} developed a cucumber disease diagnosis system and employed Grad-CAM\cite{Selvaraju2017618} to visualize the regions of interest that contributed to the model's diagnostic decisions. Although their model achieved a high performance, they observed that it occasionally responded to background regions instead of the target leaf areas due to overfitting. To mitigate this issue, Saikawa et al. applied GAN-based masking to suppress background-induced overfitting, which led to improved diagnostic performance for cucumber leaf diseases~\cite{Saikawa20195177}. 
However, subsequent experiments using a high-resolution and large-scale dataset demonstrated that background removal had only a limited impact on enhancing diagnostic performance~\cite{shibuya2022validation}.
This suggests that domain-specific characteristics are often embedded in the background and within the leaf regions.
In addition, ROI-based approaches were confirmed to be indispensable, especially in diagnosing plant pests. This task is also fine-grained, but the affected regions are smaller and show less variation than plant diseases, which makes it more critical to reduce the influence of the background~\cite{Wayama2024}.

\subsection{Various data generation using generative models}
To enhance dataset diversity for training, data augmentation using image generation techniques has been proposed. Cap et al. introduced LeafGAN~\cite{Cap20221258}, a CycleGAN-based model~\cite{Zhu20172242} that isolates leaf regions to eliminate the influence of background information. By synthetically adding disease symptoms to healthy images and using these augmented samples for training, they were able to improve classification accuracy. Furthermore, Kanno et al. proposed a method called Productive and Pathogenic Image Generation (PPIG), which addresses the limited diversity of generated images by employing a two-stage generation process~\cite{Kanno2021554}. PPIG first generates multiple healthy images from noise and then applies an image-to-image transformation model to transfer disease symptoms onto the leaf regions of these generated healthy images. This approach has been shown to enhance diagnostic performance on unseen test data by leveraging the generated images as additional training resources.
While GANs generate images based on learned data, the diversity of the generated images is often limited when trained solely on available source domain data. This limitation makes it challenging to overcome significant domain shifts in tasks such as plant disease diagnosis. In contrast, image-generation methods based on latent diffusion can produce more diverse images by leveraging not only the learnable source domain data but also large-scale pre-trained data and text prompts. This approach holds promise for application in automatic plant disease diagnosis.

\subsection{Unsupervised domain adaptation (UDA)} 
Representative methods for domain adaptation include adversarial learning-based approaches, such as Domain-Adversarial Neural Network (DANN)~\cite{Ganin2017189} and Adversarial Discriminative Domain Adaptation (ADDA)~\cite{Tzeng20172962}, which extract domain-invariant features through adversarial training, and Maximum Mean Discrepancy (MMD)-based methods~\cite{Saito20183723}, which align the distributions of the source and target domains by matching their class distributions.
In the context of plant disease diagnosis, an unsupervised domain adaptation method proposed by Wu et al.~\cite{Wu2023} has demonstrated excellent results. This method captures diverse features of disease lesions by preserving both detailed lesion information and the overall features of the leaf. It effectively mitigates domain discrepancies while achieving semantic alignment at the class level.
Additionally, a recent approach to cross-species plant disease diagnosis, proposed by Yan et al.~\cite{Yan20232555}, introduces a deep transfer learning framework for adapting mixed subdomains. This method addresses the challenge of transferring knowledge between poorly correlated domains, which is often overlooked in traditional transfer learning. 
These UDA methods are based on a typical transductive learning framework, where labels for the target domain are unavailable, but image data from the target domain can still be observed. As an alternative, pseudo-labels predicted by a model trained on the source domain are utilized. However, when there is a significant domain shift, the accuracy of these pseudo-labels becomes a concern.



\section{Target-Aware Metric Learning with Prioritized Sampling (TMPS)}

We propose that TMPS can be applied to tasks with large domain gaps when even a small number of target domain data (i.e., test environment images) are available.
TMPS is a practical learning method based on metric learning that compares training images.
Introducing a new parameter that determines the extent to which test environment images are incorporated into metric learning can significantly increase its effectiveness on limited target domain information.

The entire dataset $X$ used for training consists of a large set of source domain images $X^s$ labeled into $c$ classes and a small set of target domain images $X^t$ for each $c$ class. TMPS adopts the concept of metric learning, aiming to shorten the Euclidean distance in the feature space, which is the lower-dimensional representation of data with the same label, and to increase the distance between data with different labels in the feature space. We compare the distance between the input image $\boldsymbol{x}$ $(\boldsymbol{x}\in X)$ and the image $\boldsymbol{x_i}$ $(i\in\{1, \dots, c\})$ sampled from each class. Note that each data is converted to a low-dimensional representation through a feature extractor $f$ to calculate the distance in the feature space. Following~\cite{Hoffer2017}, the embedded similarity distribution is computed from the Euclidean distance between the features of the input image $\boldsymbol{x}$ and the compared data $\boldsymbol{x_i}$.

\input{equations/eq1}The vector $P(\boldsymbol{x};\boldsymbol{x_1},..., \boldsymbol{x_c})$ represents the similarity of $\boldsymbol{x}$ to the representative examples $\boldsymbol{x_1},\dots, \boldsymbol{x_c}$ of each class as a probability distribution. By calculating the cross-entropy loss based on the obtained embedding similarity distribution and the one-hot representation $I(\boldsymbol{x})$ of the label information indicating which class the input image $\boldsymbol{x}$ belongs to, we derive the loss $L$ based on the distance to the comparison data for each class.

\input{equations/eq2}In metric learning, the calculated loss $L$ is added as a constraint to the loss function of the original machine learning model. This encourages the model to learn embedding representations where data from the same class have similar representations while data from different classes are represented distinctly.

In order to handle test environment images more efficiently when data of the target domain is scarce, the target domain data selection probability (i.e., test field data selection probability) $p$ $(p \in [0,1])$ is introduced as a hyperparameter that determines whether the test environment images are incorporated into metric learning. Setting $p$ high increases the probability that a test environment image is sampled in metric learning, so even a small number of test environment images can contribute to adapting the embedding space. According to Eq. (\ref{equation3}), the comparison data $\boldsymbol{x_i}$ for each class is sampled from a small number of target domain images $X^t$ or a large number of source domain images $X^s$ based on the set target domain data selection probability.

\input{equations/eq3}This is expected to result in a feature space where the distance between the source and target domain data is strongly considered and domain gaps are suppressed.

\section{Experiments}

\subsection{Dataset}
\input{images/tex/leaf}

In this study, we utilized a total of 223,073 leaf surface images representing 30 disease classes, including the healthy (HE) category, for three crop types: cucumber, tomato, and eggplant. These images were collected from 23 fields and annotated by experts. 
The dataset was divided into training (source) and test (target) sets, each collected from different fields to ensure rigorous evaluation and avoid data leakage, a common issue in many existing studies. 
For the few disease classes where this separation was not feasible, we included only data collected during completely different seasons to maintain independence between the training and evaluation datasets.
To illustrate the impact of domain shift between source and target datasets, we provide examples of healthy and gray mold leaf images from both the source and target domains in Fig.~\ref{fig:domain_shift_examples}. 
It includes single large leaves or multiple leaves in the center, with varying disease symptoms and non-uniform backgrounds. 
Details of the dataset composition are provided in Table~\ref{tab:dataset}. 



\subsection{Experiment Details}
In this study, we evaluated the diagnostic performance of our proposed method, TMPS, against four comparative methods. For methods requiring target domain images, we used $10$ randomly selected labeled samples per disease category for training. 
These specific target domain images used for training were explicitly excluded from the target domain evaluation set to prevent data leakage.
The comparison methods were defined as follows:
\begin{enumerate}
\item Baseline: A classifier trained solely on the source data.
\item Metric: A classifier trained from source and target data with conventional metric learning (without prioritized sampling)~\cite{Hoffer2017}.
\item All-Train: A classifier trained on a combined dataset of both source and target domain images.
\item Fine-Tuned: A classifier initially trained on the source domain (Baseline), then fine-tuned on the target domain using only its fully connected layer.
\item TMPS: A classifier trained using the proposed TMPS method.
\end{enumerate}
The diagnostic performance of All-Train, Fine-Tuned, and TMPS were averaged over five runs.

EfficientNetV2-S~\cite{Tan202110096}, pre-trained on ImageNet-1K~\cite{Russakovsky2015211}, served as the base model for all classifiers.
Input image dimensions were resized to 512 $\times$ 512 pixels.
We employed only basic data augmentation techniques, including random cropping within 80\%--100\% of the image size, random horizontal and vertical flipping, and 90-degree rotations.

\input{tables/datasets}
\section{Results}
Table \ref{tab:result} presents the diagnostic performance of the proposed TMPS method compared with four methods across all disease classes, evaluated using the F1-score.
The proposed TMPS method demonstrates enhanced diagnostic performance across a wide range of diseases, surpassing all comparison methods.

Figure \ref{fig:result1} illustrates the relationship between discrimination performance (F1-score) and the target domain selection probability $p$, which determines the extent to which target domain images are applied in metric learning.
The results show that increasing $p$ improves discrimination performance, with the F1-score reaching its maximum at $p=0.7$ for all three crop image sets.
Note that metric learning without prioritized sampling corresponds to a scenario where the probability of target domain references is extremely low (e.g, $p < 0.02$).

\input{tables/result}
\input{images/tex/p}
\section{Discussion}
\subsection{Impact of Limited Target Domain Data Performance}
The Baseline model, which does not utilize any information from the target domain, often fails to correctly diagnose certain diseases.
This significant drop in performance highlights the serious impact of domain gaps on plant disease diagnosis in real-world settings.
Conversely, methods that incorporate even a small amount of target domain data, such as All-Train, Fine-Tuned, and TMPS, show significantly improved diagnostic performance, achieving results comparable to those of other diseases.
This confirms that incorporating even a limited amount of target domain data is crucial for practical diagnostic performance when substantial domain shifts exist.
However, the Metric method does not consistently outperform the Baseline.
While it shows improvements on some crops, it also leads to performance drops on another.
This suggests that under severe domain shifts, conventional metric learning without appropriate domain-aware mechanisms may lead to overfitting to source-specific features, thereby hindering generalization to the target domain.

\subsection{Performance Trends with Target Domain Selection Probability ($p$)}
We further analyzed how the diagnostic performance (F1-score) changes with the target domain selection probability $p$, a key parameter in TMPS that controls the extent to which target domain images are used during metric learning.
As discussed, conventional metric learning applied without specific prioritization can perform poorly in the presence of severe domain gaps.
In contrast, results consistently show that increasing $p$ in TMPS improves discrimination performance across all three crop image sets, with the F1-Score reaching its maximum at $p=0.7$ for all crops.
This highlights the substantial contribution of our prioritized sampling strategy.
A higher emphasis on target domain samples during metric learning more effectively reduces domain gaps by strategically aligning feature space distributions between source and target.
However, no improvement was observed when $p$ was increased above 0.7 for any crop.
This suggests that excessively high values of $p$ may lead to overfitting by disproportionately expanding the influence of the scarce and less diverse target field data, thereby limiting further generalization.
These findings provide empirical insight into how prioritized sampling affects the trade-off between leveraging limited target data and maintaining model robustness.
They confirm the central role of the sampling probability $p$ in the TMPS framework.

\subsection{Limitations and Future Work}
The optimal value of $p=0.7$ was empirically chosen and may vary with dataset characteristics and domain gap size.
Future work will explore theoretical or adaptive methods to determine $p$ more robustly.
We also plan to apply TMPS to other fine-grained tasks with large domain shifts, such as medical image diagnosis, where labeled data is scarce.
\section{CONCLUSIONS}
In highly challenging machine learning tasks characterized by fine-grained distinctions and significant domain gaps, the scenario where only a small amount of labeled data is available from the target domain is often realistic. The proposed TMPS was shown to have a substantial impact on the plant disease diagnosis task, demonstrating its effectiveness in such scenarios. The simple and highly versatile training strategy of TMPS is expected to yield strong results across tasks with large domain gaps.




\section*{APPENDIX}

The correspondence between the names of the plant diseases used in this experiment and the labels is as follows:
 Powdery Mildew (PM), Gray Mold (GM), Anthracnose (ANT), Cercospora Leaf Mold (CLM), Leaf Mold (LM), Late Blight (LB), Downy Mildew (DM), Corynespora Leaf Spot (CLS), Corynespora Target Spot (CTS), Leaf Spot (LS), Gummy Stem Blight (GSB), Verticillium Wilt (VW), Bacterial Wilt (BW), Bacterial Spot (BS), Bacterial Canker (BC), Cucurbit Chlorotic Yellows Virus (CCYV), Mosaic Diseases (MD), Melon Yellow Spot Virus (MYSV), Tomato Mosaic Virus (ToMV), Tomato Chlorosis Virus (ToCV), Yellow Leaf Curl (YLC), and Healthy (HE).

\section*{ACKNOWLEDGMENT}

This work was supported by the Ministry of Agriculture, Forestry, and Fisheries (MAFF), Japan, under the commissioned project study “Development of Pest Diagnosis Technology Using AI” (JP17935051), and by the Cabinet Office through the Public/Private R\&D Investment Strategic Expansion Program (PRISM).



\bibliographystyle{IEEEtran} 
\bibliography{references}

\end{document}

%% file: equations/eq1.tex
\begin{equation}
    P(\boldsymbol{x}; \boldsymbol{x_1}, \dots, \boldsymbol{x_c})_i = \frac{e^{-\| f(\boldsymbol{x}) - f(\boldsymbol{x_i}) \|^2}}{\displaystyle \sum_{j=1}^{c} e^{-\| f(\boldsymbol{x}) - f(\boldsymbol{x_j}) \|^2}}, i \in \{1 \dots c\}.
\end{equation}

%% file: equations/eq2.tex
\begin{equation}
    L(\boldsymbol{x}, \boldsymbol{x_1}, \dots, \boldsymbol{x_c}) = H_{CE}\left(I(\boldsymbol{x}), P(\boldsymbol{x}; \boldsymbol{x_1}, \dots, \boldsymbol{x_c})\right).
\end{equation}

%% file: equations/eq3.tex
\begin{equation}
    \boldsymbol{x_i} =
        \begin{cases}
            \boldsymbol{x^t_i}, & \text{with probability } p, \text{ where } \boldsymbol{x^t_i} \in X^t \\
            \boldsymbol{x^s_i}, & \text{with probability } 1-p, \text{ where } \boldsymbol{x^s_i} \in X^s. \label{equation3}
        \end{cases}
\end{equation}

%% file: images/tex/leaf.tex
\begin{figure*}[t]
  \centering
  \begin{subfigure}[b]{0.24\textwidth}
    \includegraphics[width=\linewidth]{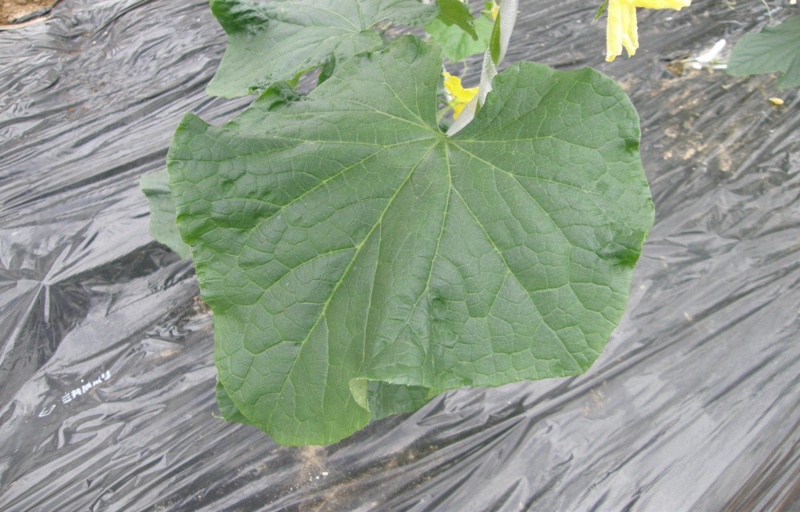}
    \caption{Healthy (source)}
    \label{fig:healthy_source}
  \end{subfigure}
  \hfill
  \begin{subfigure}[b]{0.24\textwidth}
    \includegraphics[width=\linewidth]{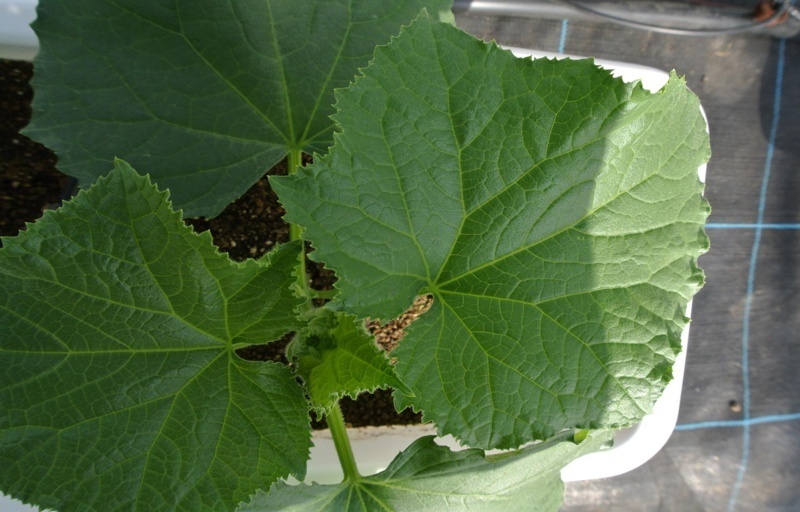}
    \caption{Healthy (target)}
    \label{fig:healthy_target}
  \end{subfigure}
  \hfill
  \begin{subfigure}[b]{0.24\textwidth}
    \includegraphics[width=\linewidth]{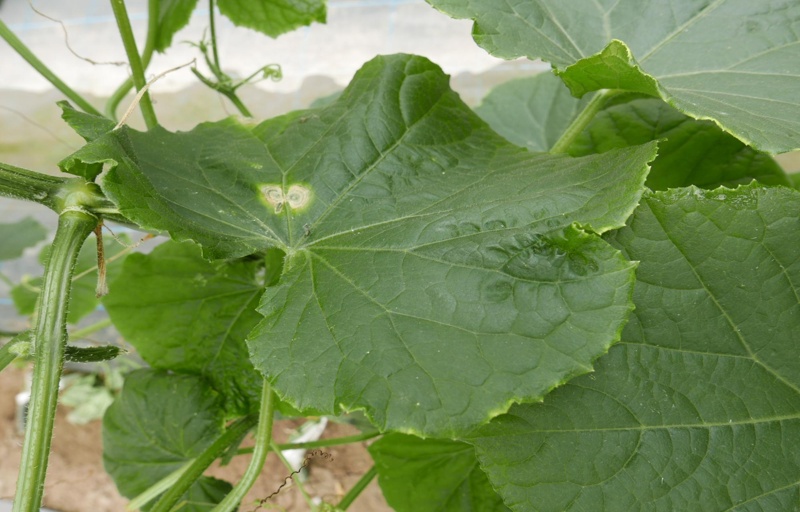}
    \caption{Gray mold (source)}
    \label{fig:graymold_source}
  \end{subfigure}
  \hfill
  \begin{subfigure}[b]{0.24\textwidth}
    \includegraphics[width=\linewidth]{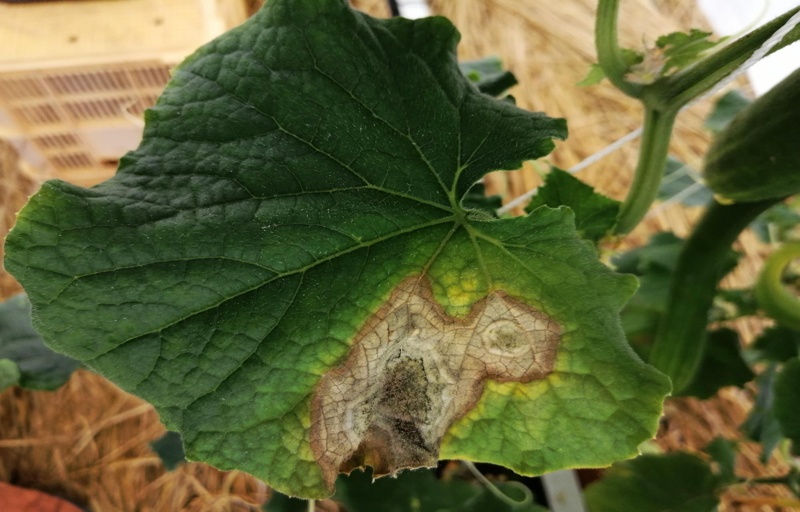}
    \caption{Gray mold (target)}
    \label{fig:graymold_target}
  \end{subfigure}
  \caption{Example images illustrating the domain shift between the source and target datasets. The images show differences in leaf appearance, background complexity, and disease symptom expression.}
  \label{fig:domain_shift_examples}
\end{figure*}

%% file: tables/datasets.tex
\begin{table}[!t]
\small
\centering
\caption{Detail of the datasets. The disease labels use abbreviations for each disease. We provide the full names of the diseases in the appendix.}
\begin{tabular}{clrr}
\toprule
\multicolumn{2}{c}{Disease} & \multicolumn{1}{c}{Source} & \multicolumn{1}{c}{Target} \\ \midrule
\multirow{12}{*}{Cucumber} & HE & 16,023                     & 5,576                      \\
                          & PM            & 7,764  & 1,898  \\
                          & GM            & 643    & 167    \\
                          & ANT           & 3,038  & 77     \\
                          & DM            & 6,953  & 2,579  \\
                          & CLS           & 7,565  & 1,813  \\
                          & GSB           & 1,483  & 374    \\
                          & BS          & 4,362  & 2,648  \\
                          & CCYV  & 5,969  & 179    \\
                          & MD            & 26,861 & 1,676  \\
                          & MYSV          & 17,239 & 1,004  \\ \cmidrule{2-4} 
                          & Total                & 97,900 & 17,991 \\ \midrule
\multirow{13}{*}{Tomato}  & HE            & 8,120   & 2,994   \\
                          & PM            & 4,490   & 4,250   \\
                          & GM            & 9,327   & 571    \\
                          & CLM           & 4,078   & 1,809   \\
                          & LM            & 2,761   & 151    \\
                          & LB            & 2,049   & 808    \\
                          & CTS          & 1,732   & 1,350   \\
                          & BW            & 2,259   & 412    \\
                          & BC            & 4,369   & 128    \\
                          & ToMV          & 3,453   & 49     \\
                          & ToCV          & 4,320   & 871    \\
                          & YLC                 & 4,513   & 1,746   \\ \cmidrule{2-4} 
                          & Total                                   & 51,471  & 15,139  \\ \midrule
\multirow{8}{*}{Eggplant} & HE                         & 12,431  & 1,122   \\
                          & PM                     & 7,936   & 938    \\
                          & GM                          & 1,024   & 166    \\
                          & LM                          & 3,188   & 732    \\
                          & LS                          & 5,510   & 118    \\
                          & VW                  & 3,176   & 354    \\
                          & BW                     & 3,415   & 462    \\ \cmidrule{2-4}
                          & Total                                   & 36,680  & 3,892   \\ \bottomrule
\end{tabular}
\noindent
\label{tab:dataset}
\end{table}

%% file: tables/result.tex
\begin{table}[t]
\small
\centering
\caption{Comparison of diagnostic capabilities of the learning methods for each disease. Note that the test field selection probability $p$ for TMPS is set to $0.7$.}
\setlength{\tabcolsep}{3.8pt}
\begin{tabularx}{\columnwidth}{llrrrrr}
\toprule
  \multicolumn{2}{c}{\multirow{2}{*}{Disease$^\dagger$}} &
  \multicolumn{5}{c}{F1-Score [\%]} \\ \cmidrule{3-7} 
 &
   &
  \multicolumn{1}{c}{Baseline} &
  \multicolumn{1}{c}{Metric} &
  \multicolumn{1}{c}{All-Train} &
  \multicolumn{1}{c}{Fine-Tuned} &
  \multicolumn{1}{c}{TMPS} \\ \midrule
\multirow{13}{*}{(C)}      & HE   & 77.7                 & 78.8          & 76.0          & 78.3          & \textbf{79.5}    \\
                           & PM   & 69.1                 & 78.0          & \textbf{81.2} & 78.0          & 80.0             \\
                           & GM   & 3.8                  & 9.2           & 62.5          & 67.6          & \textbf{85.4}    \\
                           & ANT  & 34.6                 & 24.5          & 44.6          & 35.0          & \textbf{65.0}    \\
                           & DM   & 67.9                 & 65.1          & 82.8          & 84.4          & \textbf{86.8}    \\
                           & CLS  & 60.6                 & 59.3          & 69.8          & \textbf{81.1} & 78.7             \\
                           & GSB  & 30.5                 & 30.5          & 60.6          & 64.0          & \textbf{79.2}    \\
                           & BS   & 1.7                  & 1.4           & 56.7          & 77.0          & \textbf{78.9}    \\
                           & CCYV & 61.6                 & \textbf{80.9} & 68.5          & 59.7          & 79.3             \\
                           & MD   & 58.9                 & 46.3          & 52.1          & 58.0          & \textbf{65.9}    \\
                           & MYSV & 58.1                 & 66.3          & 58.7          & 69.5          & \textbf{70.0}    \\ \cmidrule{2-7} 
                           & Ave. & 47.7                 & 49.1          & 64.9          & 68.4          & \textbf{77.2}    \\
                           &      & \multicolumn{1}{l}{} & (+1.4)        & (+17.2)       & (+20.7)       & \textbf{(+29.5)}  \\ \midrule
\multirow{14}{*}{(T))}     & HE   & 77.2                 & 83.9          & 81.1          & 79.5          & \textbf{87.7}    \\
                           & PM   & 95.4                 & 96.2          & 95.5          & 95.6          & \textbf{96.4}    \\
                           & GM   & 55.0                 & 63.2          & 69.3          & \textbf{84.8} & 75.9             \\
                           & CLM  & 86.3                 & 89.3          & 89.9          & 91.9          & \textbf{93.3}    \\
                           & LM   & 34.7                 & 38.5          & 43.6          & 48.5          & \textbf{58.5}    \\
                           & LB   & 31.3                 & 53.2          & 69.6          & \textbf{83.0} & 71.2             \\
                           & CTS  & 87.7                 & 90.3          & 89.2          & 87.7          & \textbf{94.0}    \\
                           & BW   & 75.4                 & 72.2          & 79.5          & 75.2          & \textbf{82.7}    \\
                           & BC   & 52.2                 & 55.3          & 52.8          & 55.2          & \textbf{70.5}    \\
                           & ToMV & 3.6                  & 10.6          & 13.2          & \textbf{42.4} & 27.1             \\
                           & ToCV & 50.8                 & 87.5          & 90.1          & 88.5          & \textbf{91.4}    \\
                           & YLC  & 88.3                 & 91.1          & 91.9          & \textbf{92.4} & 91.2             \\ \cmidrule{2-7} 
                           & Ave. & 61.5                 & 69.3          & 72.1          & 77.1          & \textbf{78.3}    \\
                           &      & \multicolumn{1}{l}{} & (+7.8)        & (+10.6)       & (+15.6)       & \textbf{(+16.8)} \\ \midrule
\multirow{9}{*}{(E)}       & HE   & 82.6                 & 80.5          & 85.4          & 83.6          & \textbf{86.4}    \\
                           & PM   & 92.7                 & 91.5          & 93.0          & \textbf{95.4} & 94.2             \\
                           & GM   & 70.0                 & 59.6          & 81.8          & \textbf{87.5} & 84.4             \\
                           & LM   & 89.4                 & 85.5          & 92.7          & 92.8          & \textbf{93.7}    \\
                           & LS   & 71.8                 & 56.5          & 80.7          & 79.9          & \textbf{84.5}    \\
                           & VW   & 64.9                 & 69.4          & 74.9          & 78.2          & \textbf{80.2}    \\
                           & BW   & 57.4                 & 54.3          & 64.5          & \textbf{73.7} & 72.9             \\ \cmidrule{2-7} 
                           & Ave. & 75.5                 & 71.0          & 81.9          & 84.5          & \textbf{85.2}    \\
                           &      & \multicolumn{1}{l}{} & (-4.5)       & (+6.4)        & (+9.0)        & \textbf{(+9.7)}  \\ \bottomrule
\end{tabularx}
\noindent
\footnotesize
$^\dagger$ (C): Cucumber, (T): Tomato, (E): Eggplant
\label{tab:result}
\end{table}

%% file: images/tex/p.tex
\begin{figure}[htbp]
    \centering
	 \includegraphics[width=.85\columnwidth]{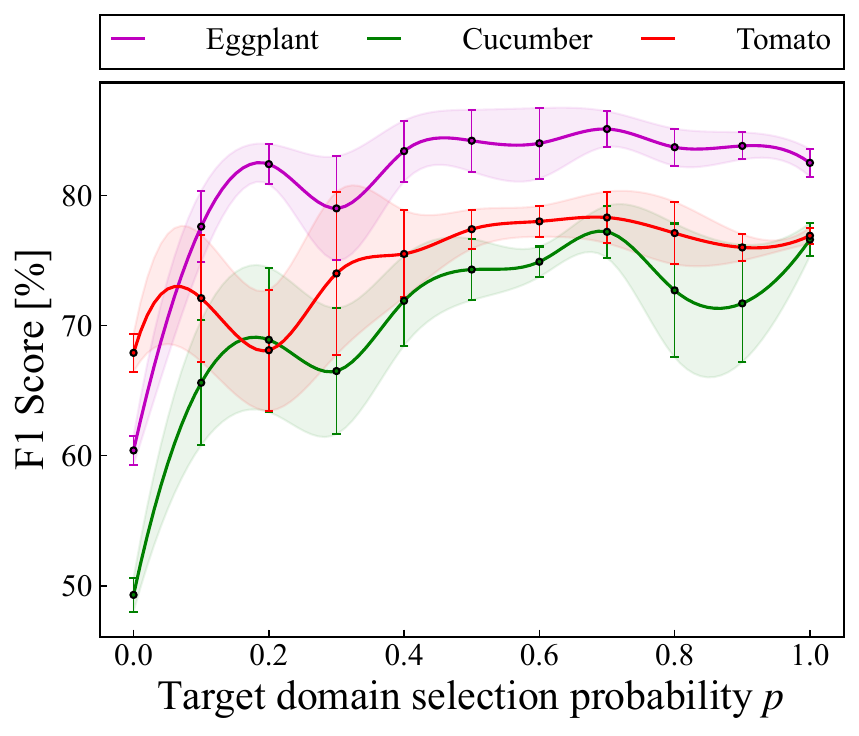}
    	 \caption{Change in performance with test field selection probability $p$. The solid line represents the average performance across five experiments, while error bars indicate the standard deviation.}
	 \label{fig:result1}
\end{figure}